\documentclass[10pt, a4paper]{article}
\usepackage{lrec}
\usepackage{multibib}
\newcites{languageresource}{Language Resources}
\usepackage{graphicx}
\usepackage{tabularx}
\usepackage{soul}

\usepackage{epstopdf}
\usepackage[latin1]{inputenc}
\usepackage{booktabs}
\usepackage[justification=centering]{caption}

\usepackage{hyperref}
\usepackage{xstring}

\usepackage{enumitem}
\setlist{nosep}
\title{Can Eye Movement Data Be Used As Ground Truth For Word Embeddings Evaluation?}

\name{Amir Bakarov}

\address{The National Research University Higher School of Economics, Moscow, Russia \\
Federal Research Center `Computer Science and Control' of Russian Academy of Sciences, Moscow, Russia \\
         amirbakarov@gmail.com\\
}

\abstract{
In recent years a certain success in the task of modeling lexical semantics was obtained with distributional semantic models. Nevertheless, the scientific community is still unaware what is the most reliable evaluation method for these models. Some researchers argue that the only possible gold standard could be obtained from neuro-cognitive resources that store information about human cognition. One of such resources is eye movement data on silent reading. The goal of this work is to test the hypothesis of whether such data could be used to evaluate distributional semantic models on different languages. We propose experiments with English and Russian eye movement datasets (Provo Corpus, GECO and Russian Sentence Corpus), word vectors (Skip-Gram models trained on national corpora and Web corpora) and word similarity datasets of Russian and English assessed by humans in order to find the existence of correlation between embeddings and eye movement data and test the hypothesis that this correlation is language independent. As a result, we found that the validity of the hypothesis being tested could be questioned. \\ \newline  }

\begin{document}

\maketitleabstract

\section{Introduction}
Dense vector word representations (\textit{word embeddings)} are gaining popularity in the scientific community because of their proved efficiency in certain downstream task. However, we are still unaware what is the most reliable evaluation method for these models. Construction of a dataset for certain downstream tasks is expensive, and some researchers argue that it is better to have a universal dataset on which can be evaluated the ability of task independent word vectors to model the structure and space of lexical semantics (or different aspects of semantics) \cite{schnabel2015evaluation}. 

Evaluation on such data is usually called \textit{intrinsic evaluation} (as an opposite to the \textit{extrinsic evaluation} which is an evaluation on downstream tasks). However, it is not actually clear how to evaluate modeling of lexical semantics. Different intrinsic evaluation tasks propose different notations of semantics, but most of the existing methods of intrinsic evaluation (like \textit{word similarity} method or \textit{word analogy} method) are criticized \cite{batchkarov2016critique,rogers2017too}. Thus, some researchers started to propose experimental evaluation techniques trying to use \textit{neuro-cognitive resources} as a gold standard for semantic modeling.

Among these techniques, there are methods of detecting neural activation patterns  on text processing (like \textit{magnetoencephalography} \cite{wehbe2014aligning}), methods of detecting reaction time on reading (like \textit{semantic priming} \cite{auguste2017evaluation}), and many others. In this work, we propose another possible data source which is \textit{eye movement on silent reading.} This data consists of measurements of time of gaze fixation on each word, amount of returns of the gaze, and so on. Eye movement data is supposed to be related to human language processing and possibly stores information about lexical semantics, so hypothetically the correlation with word embeddings could be found. Similar experiments with English eye movement data were already proposed in \cite{sogaard2016evaluating}, and the results did not show strong correlation. However, it is not clear whether the same outcome will be observed in other languages. If eye movement data is language dependent, that this could support the idea that those data are less related to the semantics. So, the \textit{main aim of this paper} is to answer the question whether word embeddings in one language correlate with eye movement data in another language (and how much). Of course, we cannot talk about investigating language independence (for that we would need many more typologically diverse languages), but possibly we could highlight some features of this comparison investigation two distinct languages.

\begin{table*}[t]
\begin{center}
\begin{tabular}{c|c|cc}
\toprule 
     \multicolumn{1}{c|}{Target word} & \multicolumn{3}{c}{The nearest neighbor word}\\
    \midrule
     & \textbf{RSC}, \textit{gaze} & \textbf{Araneum}, \textit{embeddings} & \textbf{Ruscorpora}, \textit{embeddings} \\
    \midrule
    \textbf{mikrob (\textit{microbe})} & lekarstvo (\textit{cure}) & boleznetvorniy (\textit{pathogenic}) & mikroorganizm (\textit{microorganism}) \\
    \textbf{speciya (\textit{spice})} & kuriniy (\textit{chicken}) & priprava (\textit{seasoning})  & speciya (\textit{spice}) \\
    \textbf{zanyatie (\textit{class})} & student (\textit{student}) & trenirovka (\textit{training}) &  klassniy (\textit{in class}) \\
    \textbf{plutovat (\textit{to cheat})} & ministr (\textit{minister}) & tyrit' (\textit{to steal})  & vorovat (\textit{to steal}) \\
    \textbf{chaska (\textit{cup})} & vedro (\textit{bucket}) & stakan (\textit{glass}) & chaynik (\textit{kettle}) \\
    \bottomrule
\end{tabular}
\caption{Top-1 nearest neighbors for Russian gaze vectors (first column) and Russian embeddings vectors (second and third columns). The first word is transliteration in Russian, the word in brackets is translation into English.}
\end{center}
\label{table:1}
\end{table*}

This study proposes experiments with eye movement data and word embeddings for Russian as `another' language (we consider Russian because we are native speakers of this language, so we are able to interpret the obtained results), trying to answer the question of whether word embeddings in one  language correlate with eye movement data in another, and whether the correlations between two language specific embeddings and eye movement data are comparable. Probably, distributional semantics and eye movement data process different types of semantics, so we could expect that the correlation would be low.

This paper is organized as follows. Section 2 puts our work in the context of previous studies. Section 3 describes the data that we are using in the current work. Section 4 is about the experimental setup. The results of the comparison are reported in Section 5, while Section 6 concludes the paper.

\section{Related work}

The basic idea of this work is related to the hypothesis that observable features of human text processing (like \textbf{the time of reading of a certain word}) are based not only on the surface features of a linguistic sign but also on its meaning. There are certain studies based on eye movement data that prove correlation between word semantics and reading time \cite{smith2013effect,hohenstein2013eye}. More precisely, while reading a human's brain continuously builds a model of context for already read words and integrates each new word in context, comparing it with contexts stored in the memory. The effort of this integration is inversely proportional to how probable the word is, so when an encountered word is highly unpredictable, then the time of its reading should increase.

This research is strongly inspired by \cite{sogaard2016evaluating}, which evaluated word embeddings against eye movement data from the Dundee Corpus \cite{kennedy2003dundee} using aggregate statistics of eye movement data features. Their experiments showed that there is no notable agreement between eye movement data and word embeddings. However, these experiments were performed for English only. In our work we extend the propose of \cite{sogaard2016evaluating}, making similar experiments for English and Russian to test the hypothesis whether the same outcome will be observed in other languages. 

\section{Datasets}

\subsection{Eye movement data}
Eye movement data is obtained through \textit{eye-tracking}, the process of measuring the point of the human gaze on the screen. In other words, when a person reads text on the screen, a special mechanism called eye-tracker tracks the movement of the gaze and records the information about the reading. A number of different features can be recorded, e.g. how long the gaze was fixated on a certain word, how many times the gaze returned back, and so on. We averaged the data of all examinees to obtain a feature vector for each words, so in these vectors each component would report value of each of the tracked features. We normalized these values, obtaining values of each of the features ranging from -1 to 1, and used vectors of normalized eye movement features as a gold standard for word embeddings evaluation  (here and later we will us the notion of \textbf{gaze vectors}). 

\textbf{Russian.} As a source of Russian eye movement data we used \textit{Russian Sentence Corpus (RSC)} \citelanguageresource{laurinavichyute2017russian} which contains data about reading 144 Russian sentences by 96 native speakers. After averaging examinees scores for each token and averaging word form scores for each lemma, we obtained a dataset with information on eye movements for 701 single words.

\textbf{English}. We used an English eye movement corpus which is the \textit{Provo Corpus} \citelanguageresource{luke2017provo} (the \textit{Dundee Corpus} used in \cite{sogaard2016evaluating} is not publicly available).  It contains data on reading 55 English paragraphs by 84 native speakers. We obtained vectors with information on eye movements for 1185 words from this data (with the manipulations described above).  We are also aware of another publicly available English corpus, \textit{Ghent Eye-Tracking Corpus (GECO)} \citelanguageresource{cop2017presenting} containing data on reading 5 000 English sentences by native and bilingual (for which English is a second language) English speakers (33 participants overall). In this paper the data of native speakers only is used. We obtained gaze vectors for 987 words. 

In all cases the raw data consisted of 17 logical and continuous eye movement features. The features included: 
\begin{enumerate}
\item \textit{dwell time} (summation of the duration across all fixations) on the current interest area;
\item  duration of the first fixation event that was within the current interest area;
\item dwell time of the first run within the current interest area;
\item number of all fixations in a trial falling in the first run of the current interest area;
\item total fixations falling in the interest area;
\item whether the first fixation in the interest area $N$ was preceded by a fixation in the interest area $N-1$;
\item whether the current interest area received at least one regression from the later interest areas;
\item whether regression(s) was/were made from the current interest area to the earlier interest areas;
\item dwell time from when the current interest area is first fixated until the eyes enter an interest area;
\item dwell time of the second run of fixations within the current interest area;
\item no fixation occurred in the first pass reading;
\item the duration of the first fixation made on the interest area $N+1$ after leaving the interest area $N$ in the first pass;
\item landing position in the word of the incoming saccade;
\item direction of the incoming \textit{saccade} (fast jump from one eye position to another);
\item character that was fixated by the incoming saccade;
\item whether the word was fixated once;
\item whether the word was fixated two or more times.
\end{enumerate}

In all the datasets, the recordings were obtained with an \textit{Eyelink 1000 Plus} desktop mount eye-tracker with a chin rest and a screen on which the sentences were presented.

\subsection{Word embeddings}
As a distributional model, we use \textit{Skip-Gram},
a neural predictive algorithm that updates values on the input layer (word embeddings), trying to maximize word prediction probability by minimizing the loss of the softmax function \cite{mikolov2013distributed}. The reason why we employed \textit{Word2Vec} is that it is very common in natural language processing research, evaluated and explored in many papers (note though that \cite{sogaard2016evaluating} used another type of embeddings, namely \texttt{SENNA} embeddings \cite{collobert2011natural}; however, the paper describing \texttt{SENNA} propose 4 different embedding architectures, and we are not aware which exactly architecture has been used). 

\begin{table*}[t]
\begin{center}
\begin{tabular}{l|l|ccc|ccc}
\toprule 
    & & \multicolumn{3}{c}{Similarity} & \multicolumn{3}{c}{Relatedness}\\
    \midrule
    & & \textbf{SimVerb} & \textbf{SimLex} & \textbf{WS353-Sim} &
    \textbf{MEN} & \textbf{MTurk} & \textbf{WS353-Rel} \\
    \midrule
    English & \textbf{Provo}, \textit{gaze} & 0.01 & -0.09 & -0.2 &
    -0.09 & 0.19 & 0.07 \\
    English & \textbf{GECO}, \textit{gaze} & 0.06 & -0.28 & -0.8 &
    -0.19 & 0.14 & -0.6 \\
    \midrule
    English & \textbf{Google News}, \textit{embeddings} & \boldmath$0.36^\dagger$ & $0.35^\dagger$ & \boldmath$0.77^\dagger$ &
    \boldmath$0.77^\dagger$ & $0.67^\dagger$ & \boldmath$0.64^\dagger$ \\
    English & \textbf{BNC}, \textit{embeddings} & $0.18^\dagger$ & $0.25^\dagger$ & - &
    $0.76^\dagger$ & \boldmath$0.69^\dagger$ & - \\
    \midrule
    \midrule
    Russian & \textbf{RSC}, \textit{gaze} & - & 0.24 & -0.1 & - & - & 0.05 \\
    \midrule
    Russian & \textbf{Araneum}, \textit{embeddings} & - & \boldmath$0.39^\dagger$ & $0.57^\dagger$ & - & - & $0.61^\dagger$ \\
    Russian & \textbf{Ruscorpora}, \textit{embeddings} & - & $0.28^\dagger$ & $0.74^\dagger$ & - & - & $0.57^\dagger$ \\
    \bottomrule
\end{tabular}
\caption{Performance of English word embeddings and gaze vectors across word similarity and word relatedness tasks in Spearman's correlation value. Daggers report $pval < 0.01$, absence of a symbol report $pval > 0.05$.}
\end{center}
\label{table:2}
\end{table*}

\textbf{Russian}. We used a model trained on a POS-tagged \textit{National Russian Corpus} (we will further use the term  \textbf{Ruscorpora} further) with 195 071 words in the model's vocabulary (\texttt{ruscorpora\_upos\_skipgram\_300\_5\_2018}) and a model trained on a POS-tagged \textit{Araneum Russicum Maximum} (\textbf{Areneum}) with 196 620 words in the vocabulary (\texttt{araneum\_upos\_skipgram\_300\_2\_2018}) \cite{kutuzov2016webvectors}. The models are available on \textit{Rusvectores} repository\footnote{\path{http://rusvectores.org/en/models/}}.

\textbf{English}. We used a model trained on a \textbf{Google News} corpus with 2 883 863 words in the vocabulary and a model trained on a POS-tagged British National Corpus (\textbf{BNC}) with 163 473 words in the vocabulary \cite{fares2017word}.  The models are available on \textit{Nordic Language Processing Laboratory} repository\footnote{\path{http://vectors.nlpl.eu/repository/}}. All used word vectors, both Russian and English, had the same vector dimensionality of 300. 

\subsection{Word similarity data}

\textit{Word similarity task} is the most ubiquitous technique for word embeddings evaluation. Given words $a$ and $b$, the task is to find scalar value reporting semantic distance between them. This task is strongly criticized in NLP community, and different researchers address problems like the obscurity of the notion of semantics, subjectivity of human judgments, and so on \cite{batchkarov2016critique}.

The word similarity datasets actually differ in the types of human assessments: some datasets are assessed according to semantic similarity relation (which is commonly interpreted as a synonymy, like in words \textit{mug} and \textit{cup}), while other datasets are assessed by semantic relatedness (which is interpreted as co-hyponymy, like in words \textit{cup} and \textit{coffee}).

\textbf{English.} We are aware of more than 7 datasets for word similarity available for English, but in order to propose a fair comparison with gaze vectors we need to drop words that are absent in eye movement data vocabulary. To this end, we did not used datasets like \textit{Verb-143} \cite{baker2014unsupervised}, \textit{YP-130} \cite{yang2006verb}, \textit{RG-65} \cite{rubenstein1965contextual} and \textit{MC-30} \cite{miller1991contextual}, because the amount of remaining word pairs (after dropping) was too low (lower than 5). We used the only 3 datasets with assessments by semantic similarity.  
\begin{enumerate}
\item \textit{SimVerb-3500} (234 word pairs remained for GECO, 75 word pairs remained for Provo) \cite{gerz2016simverb},
\item \textit{SimLex-999} (37 pairs remained for GECO, 41 pairs remained for Provo) \cite{hill2016simlex},
\item \textit{WordSim-353-Similarity} (WS353-Sim) (5 word pairs remained for both GECO and Provo) \cite{agirre2009study}.
\end{enumerate}

We also used 3 English datasets assessed by semantic relatedness, excluding certain datasets for the reasons described above (MTurk-287 \cite{radinsky2011word}): 

\begin{enumerate}
\item \textit{MEN} (22 pairs remained for GECO, 77 pairs remained for Provo) \cite{bruni2014multimodal},
\item \textit{MTurk-771} (7 pairs remained for GECO, 19 pairs remained for Provo) \cite{halawi2012large},
\item \textit{WordSim-353-Relatedness} (WS353-Rel) (5 pairs remained for GECO, 11 pairs remained for Provo) \cite{agirre2009study},
\end{enumerate}

\textbf{Russian}. The amount of word similarity datasets available for Russian is lower. All datasets we are aware of are translated versions of English datasets: SimLex-999, WordSim-353, RG-65 and MC-30 \cite{panchenko2016human}. The two latter were excluded from our comparison according to the reasons described above. So, we used the following datasets: 
\begin{itemize}
\item The revised version of \textit{SimLex-999}, dubbed \textit{RuSimLex-965} (21 word pairs remained) \cite{kutuzov2017size} and translated versions
\item The translated versions of \textit{WS353-Sim} (5 pairs remained) and \textit{WS353-Rel} (7 pairs remained) \cite{panchenko2016human}
\end{itemize}

\section{Experimental Setup}

To this point, we obtained two datasets with vectors (gaze vectors and word vectors) and one dataset with scalar values (human judgments on word similarity). For each word pair in each vector dataset, we computed cosine distance between the vectors corresponding to the words in a pair. Then for every word pair in each dataset we had a float in $\{0, 1\}$ reporting similarity between two words in this pair. In the end, to find correlation of the datasets we computed Spearman correlation value for distances between word pairs. The results of that comparison are presented in the following section. The code to reproduce the experiments as well links to the datasets and models are available at out GitHub\footnote{\url{https://github.com/bakarov/subconscious-embeddings/tree/master/eye-tracking/lincr2018}}. 

\section{Results and discussion}

First of all, we evaluated the gaze vector space by analyzing the nearest neighbors to certain target words. Table 1 reports the closes neighbors for Russian gaze vectors and Russian word embeddings for randomly used words (closest vectors were found with a \textit{KD-tree} neighbor search \cite{maneewongvatana1999s}). Notably, according to the notion of word relatedness, some words produced by gaze vectors seem to be very related to target words. So, we cannot say that the gaze vectors work well, but we also not conclude that they are just random.

Table 2 reports the correlation values for Russian and English gaze vectors (as well as word vectors) with word similarity assessments.  In general, the experiments show the lack of correlation even between similarity judgments and gaze vectors, giving in most cases low correlation score. Apart from the issue of embedding evaluation, this raises the problem of whether semantic similarity is a factor affecting gaze variables during reading at all. However, due to the low size of remained word pairs in pre-processed datasets, the statistical significance of obtained results could be question, so we are not able propose any confident conclusions. 

\begin{table}
\begin{center}
\begin{tabular}{c|l|cc|cc}
\toprule 
    & & \multicolumn{2}{c}{English} & \multicolumn{2}{c}{Russian}\\
    \midrule
    & & \textbf{BNC} & \textbf{GN} & \textbf{Ara} & \textbf{RC} \\
    \midrule
     English & \textbf{GECO} & \boldmath{$0.99^\dagger$} & 0.14 & - & - \\
     English & \textbf{Provo} & 0.97 & \boldmath{$0.23^\dagger$} & - & - \\
    \midrule 
     Russian & \textbf{RSC} & - & - & \boldmath{$0.65^\dagger$} & \boldmath{$0.63^\dagger$} \\
    \bottomrule
\end{tabular}
\caption{Correlation of English gaze vectors with English word embeddings and Russian gaze vectors with Russian word embeddings (values report Spearman's correlation).}
\end{center}
\label{table:4}
\end{table}

The results of pairwise comparison of distances between English gaze vectors for both eye movement datasets and embeddings vectors for both corpora are presented in Table 3 (all $pval < 0.01$). The results report very high correlation with a \textit{BNC} model and low correlation with Google News despite the fact that a \textit{Google News} model showed better results on word similarity task. On the other hand, variation in correlation scores for two English distributional models is high, while variation for Russian model is low. This fact possibly proves our hypothesis that eye movement data behaves differently for different languages. It is also interesting that the \textit{Araneum} model reports the highest correlation with gaze vectors, and it also has the best results among Russian embeddings on most of word similarity tasks, while English model show an inverse pattern.

To this end, we do not say that eye movement data is worth being used as a gold standard for evaluation since we are not actually able to interpret obtained results. It is possible that substandard embeddings (that fail on word similarity task) correlates well with eye movement data (so this data is substandard), but it is also possible that word similarity data could be substandard itself, so eye movement data detects an actually good model.  


\section{Conclusions}

In this paper, we compared word vectors, human judgments of word similarity and eye movement data of Russian and English languages. We noted that eye movement data is a some way correlated with word embeddings and even with word meaning, and the behavior of this data is different in other languages. Despite the results could be called negative, we can conclude that such data needs a more detailed investigation: that may be it would be more appropriate to use in another way of evaluation on eye movement data.

For example, one could try to train a regression model that gets word embeddings vectors and tries to predict one of the features of eye movement data. The set of words in eye movement data should be split into a train set and a test set, and an evaluation measure on test (for example, mean squared error) would report performance of word embeddings (the best embeddings should have the lowest error). The most reliable features could be selected by measuring p-value on predictions. 

So, in future we plan to make such type of evaluation, trying to only adopt one of the features instead of their vector similarity. We also want to make a more extensive comparison obtaining other eye movement datasets for other languages (like the \textit{Potsdam Corpus} \cite{stede2004potsdam}), and we plan to link other \textit{neuro-cognitive resources} (like \textit{fMRI} data) to word embeddings spaces, integrating current work in a big project about evaluation of word embeddings on different types of linguistic data.

\section*{Acknowledgements}

We would like to thank three anonymous reviewers for their valuable comments and effort to improve the manuscript. We also  thank my colleague, Andrey Kutuzov, for productive discussions on this paper.

\section{Bibliographical References}
\label{main:ref}

\bibliographystyle{lrec}
\bibliography{xample}

\begin{thebibliography}{}

\bibitem[\protect\citename{Cop \bgroup et al.\egroup }2017]{cop2017presenting}
Cop, Uschi and Dirix, Nicolas and Drieghe, Denis and Duyck, Wouter.
\newblock (2017).
\newblock {\em Presenting GECO: An eyetracking corpus of monolingual and
  bilingual sentence reading}.
\newblock Springer.

\bibitem[\protect\citename{Laurinavichyute \bgroup et al.\egroup
  }2017]{laurinavichyute2017russian}
Laurinavichyute, AK and Sekerina, Irina Alekseevna and Alexeeva, SV and
  Bagdasaryan, KA.
\newblock (2017).
\newblock {\em Russian sentence corpus: Benchmark measures of eye movements in
  Reading in cyrillic}.

\bibitem[\protect\citename{Luke and Christianson}2017]{luke2017provo}
Luke, Steven G and Christianson, Kiel.
\newblock (2017).
\newblock {\em The Provo Corpus: A large eye-tracking corpus with
  predictability norms}.
\newblock Springer.

\end{thebibliography}


\begin{thebibliography}{}

\bibitem[\protect\citename{Agirre \bgroup et al.\egroup }2009]{agirre2009study}
Agirre, E., Alfonseca, E., Hall, K., Kravalova, J., Pa{\c{s}}ca, M., and Soroa,
  A.
\newblock (2009).
\newblock A study on similarity and relatedness using distributional and
  wordnet-based approaches.
\newblock In {\em Proceedings of Human Language Technologies: The 2009 Annual
  Conference of the North American Chapter of the Association for Computational
  Linguistics}, pages 19--27. Association for Computational Linguistics.

\bibitem[\protect\citename{Auguste \bgroup et al.\egroup
  }2017]{auguste2017evaluation}
Auguste, J., Rey, A., and Favre, B.
\newblock (2017).
\newblock Evaluation of word embeddings against cognitive processes: primed
  reaction times in lexical decision and naming tasks.
\newblock In {\em Proceedings of the 2nd Workshop on Evaluating Vector Space
  Representations for NLP}, pages 21--26.

\bibitem[\protect\citename{Baker \bgroup et al.\egroup
  }2014]{baker2014unsupervised}
Baker, S., Reichart, R., and Korhonen, A.
\newblock (2014).
\newblock An unsupervised model for instance level subcategorization
  acquisition.
\newblock In {\em EMNLP}, pages 278--289.

\bibitem[\protect\citename{Batchkarov \bgroup et al.\egroup
  }2016]{batchkarov2016critique}
Batchkarov, M., Kober, T., Reffin, J., Weeds, J., and Weir, D.
\newblock (2016).
\newblock A critique of word similarity as a method for evaluating
  distributional semantic models.
\newblock In {\em Proceedings of the 1st Workshop on Evaluating Vector-Space
  Representations for NLP}, pages 7--12.

\bibitem[\protect\citename{Bruni \bgroup et al.\egroup
  }2014]{bruni2014multimodal}
Bruni, E., Tran, N.-K., and Baroni, M.
\newblock (2014).
\newblock Multimodal distributional semantics.
\newblock {\em J. Artif. Intell. Res.(JAIR)}, 49(2014):1--47.

\bibitem[\protect\citename{Collobert \bgroup et al.\egroup
  }2011]{collobert2011natural}
Collobert, R., Weston, J., Bottou, L., Karlen, M., Kavukcuoglu, K., and Kuksa,
  P.
\newblock (2011).
\newblock Natural language processing (almost) from scratch.
\newblock {\em Journal of Machine Learning Research}, 12(Aug):2493--2537.

\bibitem[\protect\citename{Fares \bgroup et al.\egroup }2017]{fares2017word}
Fares, M., Kutuzov, A., Oepen, S., and Velldal, E.
\newblock (2017).
\newblock Word vectors, reuse, and replicability: Towards a community
  repository of large-text resources.
\newblock In {\em Proceedings of the 21st Nordic Conference on Computational
  Linguistics, NoDaLiDa, 22-24 May 2017, Gothenburg, Sweden}, number 131, pages
  271--276. Link{\"o}ping University Electronic Press.

\bibitem[\protect\citename{Gerz \bgroup et al.\egroup }2016]{gerz2016simverb}
Gerz, D., Vuli{\'c}, I., Hill, F., Reichart, R., and Korhonen, A.
\newblock (2016).
\newblock Simverb-3500: A large-scale evaluation set of verb similarity.
\newblock {\em arXiv preprint arXiv:1608.00869}.

\bibitem[\protect\citename{Halawi \bgroup et al.\egroup }2012]{halawi2012large}
Halawi, G., Dror, G., Gabrilovich, E., and Koren, Y.
\newblock (2012).
\newblock Large-scale learning of word relatedness with constraints.
\newblock In {\em Proceedings of the 18th ACM SIGKDD international conference
  on Knowledge discovery and data mining}, pages 1406--1414. ACM.

\bibitem[\protect\citename{Hill \bgroup et al.\egroup }2016]{hill2016simlex}
Hill, F., Reichart, R., and Korhonen, A.
\newblock (2016).
\newblock Simlex-999: Evaluating semantic models with (genuine) similarity
  estimation.
\newblock {\em Computational Linguistics}.

\bibitem[\protect\citename{Hohenstein}2013]{hohenstein2013eye}
Hohenstein, S.
\newblock (2013).
\newblock {\em Eye movements and processing of semantic information in the
  parafovea during reading}.
\newblock {Ph.D.} thesis, Universit{\"a}tsbibliothek der Universit{\"a}t
  Potsdam.

\bibitem[\protect\citename{Kennedy \bgroup et al.\egroup
  }2003]{kennedy2003dundee}
Kennedy, A., Hill, R., and Pynte, J.
\newblock (2003).
\newblock The dundee corpus.
\newblock In {\em Proceedings of the 12th European conference on eye movement}.

\bibitem[\protect\citename{Kutuzov and Kunilovskaya}2017]{kutuzov2017size}
Kutuzov, A. and Kunilovskaya, M.
\newblock (2017).
\newblock Size vs. structure in training corpora for word embedding models:
  Araneum russicum maximum and russian national corpus.
\newblock In {\em International Conference on Analysis of Images, Social
  Networks and Texts}, pages 47--58. Springer.

\bibitem[\protect\citename{Kutuzov and Kuzmenko}2016]{kutuzov2016webvectors}
Kutuzov, A. and Kuzmenko, E.
\newblock (2016).
\newblock Webvectors: A toolkit for building web interfaces for vector semantic
  models.
\newblock In {\em International Conference on Analysis of Images, Social
  Networks and Texts}, pages 155--161. Springer.

\bibitem[\protect\citename{Maneewongvatana and
  Mount}1999]{maneewongvatana1999s}
Maneewongvatana, S. and Mount, D.~M.
\newblock (1999).
\newblock Its okay to be skinny, if your friends are fat.
\newblock In {\em Center for Geometric Computing 4th Annual Workshop on
  Computational Geometry}, volume~2, pages 1--8.

\bibitem[\protect\citename{Mikolov \bgroup et al.\egroup
  }2013]{mikolov2013distributed}
Mikolov, T., Sutskever, I., Chen, K., Corrado, G.~S., and Dean, J.
\newblock (2013).
\newblock Distributed representations of words and phrases and their
  compositionality.
\newblock In {\em Advances in neural information processing systems}, pages
  3111--3119.

\bibitem[\protect\citename{Miller and Charles}1991]{miller1991contextual}
Miller, G.~A. and Charles, W.~G.
\newblock (1991).
\newblock Contextual correlates of semantic similarity.
\newblock {\em Language and cognitive processes}, 6(1):1--28.

\bibitem[\protect\citename{Panchenko \bgroup et al.\egroup
  }2016]{panchenko2016human}
Panchenko, A., Ustalov, D., Arefyev, N., Paperno, D., Konstantinova, N.,
  Loukachevitch, N., and Biemann, C.
\newblock (2016).
\newblock Human and machine judgements for russian semantic relatedness.
\newblock In {\em International Conference on Analysis of Images, Social
  Networks and Texts}, pages 221--235. Springer.

\bibitem[\protect\citename{Radinsky \bgroup et al.\egroup
  }2011]{radinsky2011word}
Radinsky, K., Agichtein, E., Gabrilovich, E., and Markovitch, S.
\newblock (2011).
\newblock A word at a time: computing word relatedness using temporal semantic
  analysis.
\newblock In {\em Proceedings of the 20th international conference on World
  wide web}, pages 337--346. ACM.

\bibitem[\protect\citename{Rogers \bgroup et al.\egroup }2017]{rogers2017too}
Rogers, A., Drozd, A., and Li, B.
\newblock (2017).
\newblock The (too many) problems of analogical reasoning with word vectors.
\newblock In {\em Proceedings of the 6th Joint Conference on Lexical and
  Computational Semantics (* SEM 2017)}, pages 135--148.

\bibitem[\protect\citename{Rubenstein and
  Goodenough}1965]{rubenstein1965contextual}
Rubenstein, H. and Goodenough, J.~B.
\newblock (1965).
\newblock Contextual correlates of synonymy.
\newblock {\em Communications of the ACM}, 8(10):627--633.

\bibitem[\protect\citename{Schnabel \bgroup et al.\egroup
  }2015]{schnabel2015evaluation}
Schnabel, T., Labutov, I., Mimno, D.~M., and Joachims, T.
\newblock (2015).
\newblock Evaluation methods for unsupervised word embeddings.
\newblock In {\em EMNLP}, pages 298--307.

\bibitem[\protect\citename{Smith and Levy}2013]{smith2013effect}
Smith, N.~J. and Levy, R.
\newblock (2013).
\newblock The effect of word predictability on reading time is logarithmic.
\newblock {\em Cognition}, 128(3):302--319.

\bibitem[\protect\citename{S{\o}gaard}2016]{sogaard2016evaluating}
S{\o}gaard, A.
\newblock (2016).
\newblock Evaluating word embeddings with fmri and eye-tracking.
\newblock {\em ACL 2016}, page 116.

\bibitem[\protect\citename{Stede}2004]{stede2004potsdam}
Stede, M.
\newblock (2004).
\newblock The potsdam commentary corpus.
\newblock In {\em Proceedings of the 2004 ACL Workshop on Discourse
  Annotation}, pages 96--102. Association for Computational Linguistics.

\bibitem[\protect\citename{Wehbe \bgroup et al.\egroup
  }2014]{wehbe2014aligning}
Wehbe, L., Vaswani, A., Knight, K., and Mitchell, T.
\newblock (2014).
\newblock Aligning context-based statistical models of language with brain
  activity during reading.
\newblock In {\em Proceedings of the 2014 Conference on Empirical Methods in
  Natural Language Processing (EMNLP)}, pages 233--243.

\bibitem[\protect\citename{Yang and Powers}2006]{yang2006verb}
Yang, D. and Powers, D.~M.
\newblock (2006).
\newblock Verb similarity on the taxonomy of wordnet.
\newblock In {\em The Third International WordNet Conference: GWC 2006}.
  Masaryk University.

\end{thebibliography}

\section{Language Resource References}
\label{lr:ref}
\bibliographystylelanguageresource{lrec}
\bibliographylanguageresource{xample}

\end{document}